\documentclass{article}

\usepackage{arxiv}
\usepackage[numbers]{natbib}
\usepackage[utf8]{inputenc} 
\usepackage[T1]{fontenc}    
\usepackage{url}            
\usepackage{booktabs}       
\usepackage{amsfonts}       
\usepackage{nicefrac}       
\usepackage{microtype}      
\usepackage{lipsum}		
\usepackage{graphicx}
\usepackage{doi}
\usepackage{amsmath}
\usepackage{xcolor}         
\usepackage[verbose=true,letterpaper]{geometry}
\usepackage{algorithmic}
\usepackage{textcomp}
\usepackage{framed,multirow}

\usepackage{amssymb}
\usepackage{latexsym}
\usepackage{array}

\title{Unsupervised Segmentation of Fetal Brain MRI using Deep Learning Cascaded Registration}

\author{\href{https://orcid.org/0009-0001-7512-0256}{\includegraphics[scale=0.06]{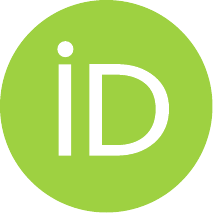}\hspace{1mm}Valentin Comte}$^1$ \and Mireia Alenya$^1$ \and Andrea Urru$^1$ \and Judith Recober$^1$ \and Ayako Nakaki$^{2,4}$ \and Francesca Crovetto$^{2,4}$ \and Oscar Camara$^1$ \and Eduard Gratacós$^{2,4,5}$ \and Elisenda Eixarch$^{2,4,5}$ \and Fàtima Crispi$^{2,4,5}$ \and Gemma Piella$^1$ \and Mario Ceresa$^{1,6}$ \and Miguel A. González Ballester$^{1,3}$}

\begin{document}

\date{}
\maketitle

{\small$^1$ BCN MedTech, Department of Information and Communication Technologies, Universitat Pompeu Fabra, Barcelona, Spain\\
$^2$  Maternal Fetal Medicine, BCNatal, Center for Maternal Fetal and Neonatal Medicine, Barcelona, Spain\\
$^3$ Institución Catalana de Investigación y Estudios Avanzados (ICREA), Barcelona, Spain\\
$^4$ Institut d’Investigacions Biomèdiques August Pi i Sunyer (IDIBAPS), Barcelona, Spain\\
$^5$ Centre for Biomedical Research on Rare Diseases (CIBERER), Barcelona, Spain\\
$^6$ European Commission, Joint Research Centre (JRC), Ispra, Italy\\}

\begin{abstract}
Accurate segmentation of fetal brain magnetic resonance images is crucial for analyzing fetal brain development and detecting potential neurodevelopmental abnormalities. Traditional deep learning-based automatic segmentation, although effective, requires extensive training data with ground-truth labels, typically produced by clinicians through a time-consuming annotation process. To overcome this challenge, we propose a novel unsupervised segmentation method based on multi-atlas segmentation, that accurately segments multiple tissues without relying on labeled data for training. Our method employs a cascaded deep learning network for 3D image registration, which computes small, incremental deformations to the moving image to align it precisely with the fixed image. This cascaded network can then be used to register multiple annotated images with the image to be segmented, and combine the propagated labels to form a refined segmentation. Our experiments demonstrate that the proposed cascaded architecture outperforms the state-of-the-art registration methods that were tested. Furthermore, the derived segmentation method achieves similar performance and inference time to nnU-Net while only using a small subset of annotated data for the multi-atlas segmentation task and none for training the network. Our pipeline for registration and multi-atlas segmentation is publicly available at \url{https://github.com/ValBcn/CasReg}.
\end{abstract}


\section{Introduction}

Fetal brain Magnetic Resonance Imaging (MRI) is an important asset in the early diagnosis of neurodevelopment abnormalities, such as those derived from Intrauterine Growth Restriction, Corpus Callosum Agenesis, or Ventriculomegaly \cite{benkarim2018cortical, benkarim2020novel, hahner2019global, torrents2019segmentation, benkarim2017toward}. In comparison with the adult brain, acquiring fetal brain MRI is a challenging procedure both at a clinical and technical level. A major difficulty is the appearance of motion artifacts caused by the fetal movements and the mother's breathing. To tackle this issue, snapshot imaging techniques, such as T2-weighted Single Shot Fast Spin Echo, are preferred to 3D volumetric imaging. These techniques produce low-resolution images with minimized motion artifacts that can be acquired along all planes. 3D Super-resolution reconstruction algorithms, such as proposed by \cite{ebner2020automated} or \cite{kuklisova2012reconstruction}, are then applied to combine the low resolution images and form high resolution 3D volumes.

Once the high resolution fetal MRI are reconstructed, the crucial first step in the analysis of fetal brain morphology is tissue segmentation, as it allows identifying specific regions of interest, such as the cortex, white matter, ventricles, or thalamus, for instance, that may be affected in cases of abnormal brain development. In recent years, numerous deep learning (DL) approaches for medical image segmentation have been proposed, many of which utilize convolutional neural networks with an encoder-decoder structure, such as U-Net \cite{ronneberger2015u}. These approaches take advantage of the deconvolution concept \cite{zeiler2014visualizing} and incorporate skip connections between layers of the same dimension on either side of the network in order to preserve fine details in the feature maps. Before DL techniques became the dominant method for image segmentation, multi-atlas segmentation (MAS) was often regarded as the most accurate approach \cite{iglesias2015multi}.

MAS is a prevalent technique in medical image analysis, providing enhanced spatial and topological consistency in comparison to automatic segmentation approaches. MAS employs multiple atlases, or reference images, which are registered to the target image to guide the segmentation process. The quality of the segmentation with MAS primarily depends on the registration accuracy, but also on the atlas selection for a specific task. Generally, segmentation accuracy increases with the number of atlases, but this comes with a trade-off of increased computational demands.

Classical registration methods use intensity-based similarity metrics, such as cross-correlation, mutual information, or sum of square distance, to measure the degree of alignment between the fixed image and the warped moving image, and maximize the similarity metric by iterative optimization methods. These methods are effective, but can be computationally intensive and time-consuming. In recent years, DL-based registration methods have been proposed as a faster alternative to classical methods, achieving similar results in a shorter amount of time. These DL-based approaches are typically based on UNet-like networks and use the same intensity-based similarity metrics as terms in the loss function. However, the application of DL-based registration in MAS has not been thoroughly investigated.

In this work, we propose a MAS approach based on unsupervised DL cascaded registration, which includes:

\begin{itemize}
    \item A novel DL-based image registration method that utilizes cascaded networks to predict a series of deformation fields. These deformation fields are then combined to align accurately the moving image with the fixed image.
    \item A contracted architecture of VoxelMorph \cite{Voxelmorph} for the cascaded networks, which reduces the memory cost while achieving similar results. This allows using more cascades, improving the overall performances.
    \item A MAS method using our cascaded registration model, which obtains similar results to nnU-Net \cite{isensee2021nnu}, without using any labelled data for training.
\end{itemize}

The remaining part of this article is structured as follows. Section \ref{related_works} reviews works related to MAS and DL registration methods. Section \ref{method} first presents the proposed cascaded registration model, then describes the derived MAS method. In Section \ref{experimental_settings} we discuss the experimental setting of our experiments, the results of which are presented in Section \ref{results}. Finally, we conclude and discuss further works and improvements in Section \ref{conclusion}.

\section{Related works}\label{related_works}

\subsection{Registratiom}

In the past few years, several DL-based methods for medical image registration have been proposed. Most of them take advantage of spatial transformer networks \cite{jaderberg2015spatial}, which adds a differentiable spatial transformation layer in order to produce dense deformation fields. In their work, \cite{li2018non} proposed a self-supervised method, that registers directly pairs of 3D brain MRI scans without training the network, similarly to classical registration methods. They used a Fully Convolutional Network (FCN) that updates its parameters with a loss based on normalized cross-correlation (NCC) and a total variation term for the regularization of the deformation field. In \cite{fan2018adversarial}, the authors do not use similarity metrics for the training of their CNN, but a discriminator network, that evaluates whether or not a pair of images is sufficiently aligned by the registration. Balakrishnan et al. presented VoxelMorph \cite{Voxelmorph}, a CNN autoencoder with a UNet-like architecture, featuring ReLU activation and skip connection between the feature maps. They showed the efficiency of their methods on 3D brain MRI scans, yielding superior results compared to traditional methods \cite{avants2011reproducible} in a much shorter amount of time. In their study, \cite{krebs2019learning} use a conditional variational autoencoder to learn a diffeomorphic deformation field from a latent space representation of the images. The network outputs a velocity field, which is then transformed into a deformation field using an exponentiation layer. Instead of using traditional regularization techniques, this method regularizes the spatial properties of the velocity field using a Gaussian convolution. Recursive cascaded networks for registration were first proposed by \cite{zhao2019recursive}. Their model consists of several cascaded networks that warp progressively the moving image into the fixed image. They use VoxelMorph \cite{Voxelmorph} and Volume Tweening Network (VTN) \cite{zhao2019unsupervised} as baseline networks. 

\subsection{Multi-atlas segmentation}

MAS can be divided into four main steps: image registration, atlas selection, label propagation and label fusion. During the image registration step, each atlas is aligned with the target image using a registration algorithm. This step is crucial because it allows accurate propagation of the atlases labels onto the target image. The atlas selection step involves choosing which atlases to use based on factors such as the similarity between the atlases and the target image before or after registration, and the relevance of the atlases to the specific task at hand. During the label propagation step, the labels from the selected atlases are transferred onto the target image using the deformation fields obtained from the image registration step. Finally, in the label fusion step, the propagated labels are combined to create the final segmentation of the target image.

Early atlas-based image segmentation techniques involved aligning a single atlas with the target image through the use of registration. This was due to the limited availability of manually annotated images and to the high computational cost of registration at the time. Once the atlas was registered with the target image, the resulting transformation (deformation field) was used to propagate the atlas segmentation labels onto the target image \cite{christensen1997volumetric,collins1995automatic,lancaster1997automated}. However, a single registration may not encompass all the anatomical variation from one brain to another. In the early 2000s, MAS gained popularity thanks to several publications \cite{rohlfing2004evaluation,klein2005mindboggle,heckemann2006automatic}. \cite{rohlfing2004evaluation} were the first to apply multiple registration techniques for atlas-based segmentation of three-dimensional microscopy images of bee's brains. Their work demonstrated that MAS was superior to single atlas-based segmentation. Building on this research, \cite{klein2005mindboggle} showed the effectiveness of MAS for segmenting human brain MRI scans. Their results indicated that using a larger number of atlases in MAS can help capture a wider range of anatomical variability and improve segmentation accuracy. Typically, intensity-based registration tools such as \cite{avants2011reproducible}, \cite{klein2009elastix}, or \cite{rueckert1999nonrigid} are used to compute one independent registration between each atlas and the target. Some other studies have experimented with running the registration step multiple times using different parameter settings and then combining all the resulting propagated labels, as in \cite{rohlfing2005multi}. In the context of perinatal brain segmentation, \cite{ballester2022automatic} further developed the concept of pre-computed registration by building their segmentation pipeline upon the existing dHCP pipeline \cite{makropoulos2018developing} for neonatal brain segmentation. Rather than registering the image to segment with all the atlases, their pipeline solely registers the image with a common template that has been previously registered with all the atlases. This approach requires to perform only one registration, which significantly reduces computational time. Despite all those progresses and optimizations, MAS using traditional registration algorithm remains computationally expensive and time-consuming. Alternatively, MAS using DL registration could considerably reduce the computation time while achieving similar or better results.

\begin{figure*}[!ht]
  \centering
  \includegraphics[width=.9\textwidth]{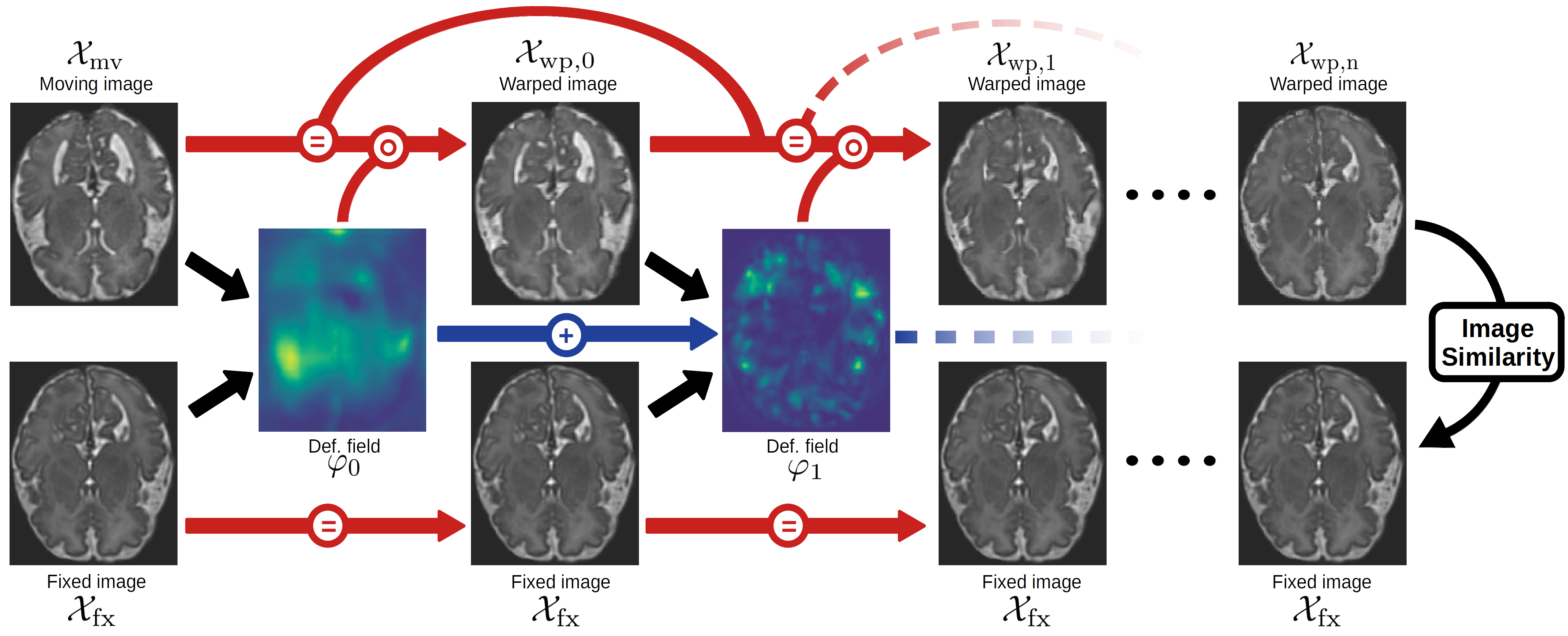}
  \caption{Overview of the cascaded registration model.}
  \label{cascade_flows}
\end{figure*}

Our proposed method improves upon existing cascaded registration techniques by utilizing a model that reduces information loss through accumulation of the deformation fields produced by the cascades. It also saves memory by employing a contracted architecture that enables to stack more cascades in order to achieve better registration results. Furthermore, we go beyond existing DL registration works by implementing a derived MAS approach. Our experiments have shown that the segmentation performances and computation time are similar to those achieved by nnU-Net \cite{isensee2021nnu}, making it a much faster alternative than the classical MAS approaches. This integrated solution offers a more comprehensive approach to medical image segmentation, and does so without the need for labelled data for training.

\section{Method}\label{method}

\subsection{Cascaded registration}\label{casc_reg_sec}

We present a cascaded DL model designed to register 3D medical images. Our model consists of cascaded networks generating successive deformation fields that are then combined to align the moving image with the fixed image. This cascaded approach allows us to divide the registration process into smaller, simpler transformations that process the images at various spatial resolutions. In mathematical terms, the process of registering two 3D images can be expressed as follows: let $\mathcal{X}_\mathrm{mv}$ and $\mathcal{X}_\mathrm{fx}$ be the moving and fixed image, of size $H \times W \times L$, defined over a 3-dimensional domain $\Omega \subset \mathbb{R}^3$. The goal of registration is to align the moving image with the fixed image using a spatial transformation $\varphi:\mathbb{R}^3 \rightarrow \mathbb{R}^3$. The warped moving image $\mathcal{X}_\mathrm{wp}$ which is generated by this transformation should be as similar as possible to the fixed image $\mathcal{X}_\mathrm{fx}$:

\begin{equation}
    \centering
    \mathcal{X}_\mathrm{wp} =  \mathcal{X}_\mathrm{mv} \circ \varphi \approx \mathcal{X}_\mathrm{fx}.
    \label{warped_img_eq}
\end{equation}

Figure \ref{cascade_flows} describes how our cascaded model operates. The first network takes the moving and fixed images, $\mathcal{X}_\mathrm{mv}$ and $\mathcal{X}_\mathrm{fx}$, as inputs and produces a dense deformation field $\varphi_0$ that partially aligns the moving image with the fixed image, forming $\mathcal{X}_\mathrm{wp,0}$. Subsequently, the second network inputs $\mathcal{X}_\mathrm{wp,0}$ and $\mathcal{X}_\mathrm{fx}$ to generate $\varphi_1$, which is added with the previous deformation field $\varphi_0$ to warp $\mathcal{X}_\mathrm{mv}$ into $\mathcal{X}_\mathrm{wp,1}$. This process is repeated with successive networks in a recursive manner, resulting in the final deformation field that fully aligns the moving image with the fixed image:

\begin{align}
\begin{split}
    \mathcal{X}_\mathrm{wp,n} = \mathcal{X}_\mathrm{mv} \circ \sum_{i=0}^n  \varphi_i  .
\end{split}
\end{align}

One major feature of our model is the cumulative nature of the cascades. Rather than successively warping the moving image, which can result in a loss of information due to the repeated interpolation operation at each step, we propose to only warp the moving image with the sum of the previous deformation fields.

For this work, we use the loss function proposed by \cite{Voxelmorph}, defined as the sum of a similarity loss, $\mathcal{L}_\mathrm{sim}$, which encourages the resemblance of the warped image with the fixed image, and a regularization loss, $\mathcal{L}_\mathrm{smooth}$, which favors smooth and reversible deformation fields. Hence, the full loss is written as:

\begin{align}
    \mathcal{L} = & ~\mathcal{L}_\mathrm{sim} + \lambda \mathcal{L}_\mathrm{smooth} \\
    = &- ~ NCC(\mathcal{X}_\mathrm{fx},\mathcal{X}_\mathrm{wp,n}) + \lambda \sum_{p \in \Omega} || \nabla \varphi(p)||^2
    \label{loss}
\end{align}

\begin{figure}[h!]
  \begin{minipage}[c]{0.6\textwidth}
    \caption{Illustration of the successive partial deformation fields produced by the cascaded networks (bottom) and the final deformation (top). The color mapping is done by translating the three components of the deformation into RGB colors. We observe the increasing spatial frequency of the deformation fields, encoding different levels of detail.} 
  \end{minipage}\hfill
  \begin{minipage}[l]{0.35\textwidth}
    \includegraphics[width=\textwidth]{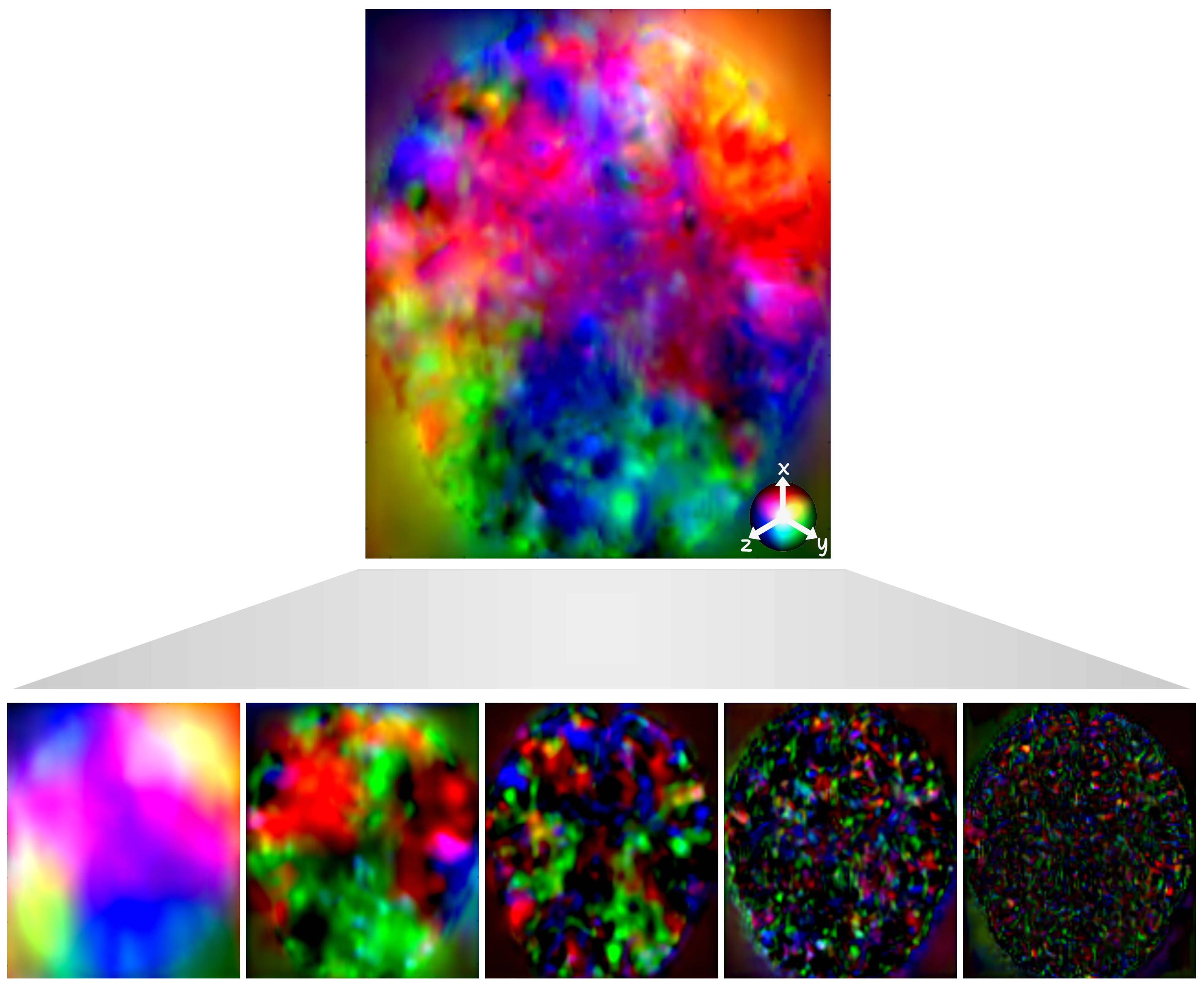}
    \label{flowz}
  \end{minipage}
\end{figure}

\noindent
where $\lambda$ is the regularization trade-off parameter. $\mathcal{L}_\mathrm{sim}$ is based on the negative NCC between the warped and fixed images, while $\mathcal{L}_\mathrm{smooth}$ represents the gradient of the predicted deformation field. Both terms of the loss functions are computed on the final outputs of the networks, $\mathcal{X}_\mathrm{wp,n}$ and $\varphi = \sum_{i=0}^n\varphi_i$  respectively, ensuring a collaborative behaviour of the cascades. We evaluate the performance of our methods using Dice score \cite{dice1945measures}, based on the segmentation of anatomical structures.
All our results are given with the associated standard error.

\subsubsection*{Contracted architecture}\label{sec_cont}

\begin{figure}[h!]
  \centering
  \includegraphics[width=.45\textwidth]{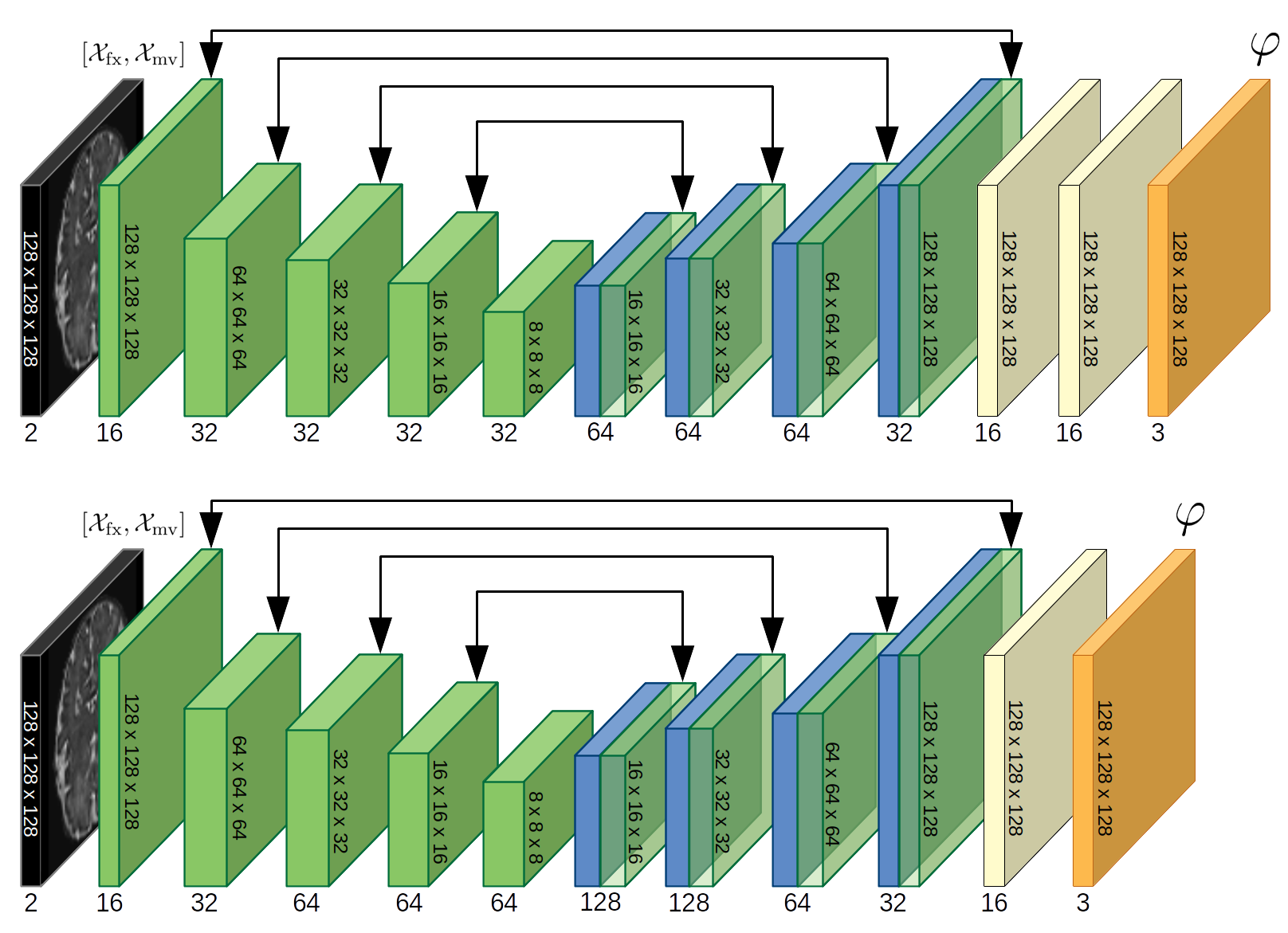}
  \caption{Original VoxelMorph architecture (top) and the proposed alternative (bottom).}
  \label{arch_alt}
\end{figure}

A challenging aspect of cascaded networks is the increasing memory cost with the number of cascades. To address this issue, we adopted an alternative architecture designed to reduce the number of learnable parameters (Figure \ref{arch_alt}). This alternative architecture presents higher number of feature maps for the hidden layers, enabling to encode more information for a marginal increase in resources, and it eliminates the final convolution layer on the full-sized image, leading to a substantial decrease in memory consumption. Those modifications enable to achieve equivalent performance while consuming fewer memory resources, as later discussed in Section \ref{sec_cont2}.

\subsection{Multi-atlas segmentation}\label{sec_seg}

Following the cascaded registration method presented in Section \ref{casc_reg_sec}, we propose a segmentation method based on multi-atlas registration. This approach involves using a series of atlas images, each associated with ground-truth labels, and registering them to the target image. The best aligned atlases are then selected based on their cross-correlation with the target image. This helps to ensure that only the most appropriate atlas images are used for the final segmentation. Finally, we propagate the labels of the selected atlases and we combine them using a local weighted voting strategy, which consists of assigning locally greater weight to labels from atlas images better aligned with the target, based on a local similarity measure.

\subsubsection*{Atlas selection}\label{atlas_selection}

The main motivations for selecting specific atlases for use in MAS are, on one hand, to use the more suitable atlases for the task at hand, discarding the ones that overly differ in terms of appearance or morphology, and that may lower the accuracy of the segmentation, and on the other hand, to reduce the computational time. The atlas selection can be performed prior to or after registration. In the case of a selection prior to registration, the atlases can be chosen based on demographic data, such as the gestational age of the fetus or pathology, for instance, or on similarity measures with the target image. Nonetheless, post-registration selection based on image similarity should be considered as a more robust approach, as it can provide an accurate evaluation of how well a particular atlas matches the fixed image. With classical methods, post-registration atlas selection was regarded as unpractical due to the large computational time required to compare each pair of images. This limitation vanishes with DL methods, as they can register images in only a fraction of a second. Different similarity measures can be used for atlas selection, such as NCC, Mean Square Error (MSE), and Structural Similarity Index Measure (SSIM). We evaluated those similarity measures and chose the most suitable one for the atlas selection in Section \ref{res_atlas_selection}.

\subsubsection*{Label fusion}\label{label_fusion}

Label fusion is the last critical phase of MAS, as the quality of the final segmentation strongly hinges on the chosen strategy. It consists of combining the propagated labels from the atlases in order to generate a refined segmentation of the target image. The most straightforward method is Majority Voting (MV) \cite{kittler1998combining}, which simply selects, for each voxel, the label that is most commonly assigned among the propagated labels. While this approach is simple and easy to implement, it does not take into account the local similarity between the transformed atlases and the target image, possibly resulting in a less accurate segmentation.

In this work, we adopt a technique called Local Weighting Voting (LWV) \cite{artaechevarria2009combination}, which assigns more weight to labels that correspond to areas of high local similarity between the transformed atlases and the target image. By taking local similarity into account, the resulting segmentation is more accurate and better capture the complex structure and variability within the target image, compared to MV. LWV assigns a weight $w$ to each voxel $i$ of the propagated label $k$ as follows:

\begin{equation}
    w_{k,i} = |m(i\in \omega)|^g
\end{equation}

\noindent
where $\omega$ is a region of size $d^3$ around the voxel $i$, $m$ is the average local similarity metrics in the region $\omega$, and $g$ is the gain factor, which can take different values depending on the similarity metrics used. 

\section{Experimental settings}\label{experimental_settings}

The model was developped using Python 3.8, Pytorch 1.11, and a GPU Nvidia Titan Xp with 12GB of memory. The baseline architecture of the cascaded networks is a contracted version of \cite{Voxelmorph}, designed to limit memory usage (see Section \ref{sec_cont}). The networks were trained for 500 epochs with 1000 iterations per epoch using the Adam optimizer, an exponentially decaying learning rate $lr_\mathrm{epoch} = 3 \cdot 10^{-4} \cdot e^{-3\mathrm{epoch}/500}$ and a batch size of $1$.

Our experiments were conducted using two fetal brain MRI datasets:

\begin{itemize}
    \item The IMPACT dataset \cite{crovetto2021effects,ballester2022automatic}: composed of 170 fetal brain MRI scans between 32 and 37 gestational weeks. The scans were aquired in Hospital San Joan de Déu and Hospital Clínic of Barcelona within the IMPACT study. Single-shot fast spin-echo T2-weighted was performed using two 3.0 T MRI scanners, Philips Ingenia (repetition/echo time: 1570/150 ms, slice thickness: 3 mm, field of view $290 \times 250$ mm, voxel spacing $0.7 \times 0.7 \times 3.0$ mm), and Siemens Magneton Vida (repetition/echo time: 1390/160 ms, slice thickness: 3 mm, field of view $230 \times 230$ mm, voxel spacing $1.2 \times 1.2 \times 3.0$ mm). The low resolution stacks were reconstructed using the \cite{ebner2020automated} super-resolution reconstruction (SRR) pipeline. All  fetuses included in this study did not have any major malformation. Every MRI scans in those datasets were annotated by clinicians with 7 anatomical labels (cerebro-spinal fluid, white matter, grey matter, ventricles, cerebellum, thalamus, and brain stem). The dataset is split into 140-10-20 for training, validation and test sets, respectively.
    
    \item The FeTa dataset \cite{payette2021automatic}: publicly available dataset composed of 160 T2-weighted fetal brain MRI scans between 20 and 35 gestational weeks, including neurotypical and pathological subjects. The scans were acquired in four institutions and the high resolution 3D volumes were obtained using two different SRR methods \cite{ebner2020automated, kuklisova2012reconstruction}. Every MRI scans in those datasets were annotated by clinicians with 7 anatomical labels (cerebro-spinal fluid, white matter, grey matter, ventricles, cerebellum, thalamus, and brain stem). The dataset is split into 130-10-20 for training, validation and test sets, respectively.
\end{itemize}

The seven anatomical labels are included for the validation and test sets (although they are not required for the validation set). As a preprocessing step, the images were cropped to remove the non-brain regions, resized to $128 \times 128 \times 128$ voxels, and normalized between 0 and 1.

\section{Results}\label{results}

\subsection{Cascaded registration}

\subsubsection*{Contracted architecture}\label{sec_cont2}

As explained in Section \ref{sec_cont}, we developed a modified version of the VoxelMorph \cite{Voxelmorph} architecture in order to reduce the memory cost and enable the use of more cascades. This modified architecture is referred to as the contracted architecture. Table \ref{contracted} presents the average Dice scores obtained with both the original Voxelmorph architecture and the contracted architecture for different numbers of cascades, as well as the GPU usage during inference. The contracted architecture, while achieving slightly worse results for the same number of cascades, allows to stack more cascades, resulting in better overall performances. We also observe from the table that the percentage of negative Jacobian values decreases with the increasing number of cascades for both architectures. This suggests that the cascades are able to decompose the transformation field into smaller and simpler transformations, more likely to be invertible.

\begin{table*}[!ht]
\centering
\begin{tabular}{>{\centering\arraybackslash}p{1.9cm}>{\centering\arraybackslash}p{2cm}>{\centering\arraybackslash}p{1.9cm}>{\centering\arraybackslash}p{1.8cm}>{\centering\arraybackslash}p{2cm}>{\centering\arraybackslash}p{1.9cm}>{\centering\arraybackslash}p{1.8cm}}
    \hline
    ~ & \multicolumn{3}{c}{Original architecture} & \multicolumn{3}{c}{Contracted Architecture}\\
    \hline
    N cascades & Dice & $\%|J_\phi | \leq 0 $ & GPU & Dice & $\%|J_\phi | \leq 0$ & GPU\\
    \hline
    1 &  0.818 $\pm$ 0.028 & 1.70 $\pm$ 0.06 & 4.6 & 0.805 $\pm$ 0.029 & 1.94 $\pm$ 0.06 & 3.3\\
    2 &  0.844 $\pm$ 0.023 & 0.76 $\pm$ 0.06 & 7.9 & 0.836 $\pm$ 0.023 & 0.83 $\pm$ 0.05 & 4.5\\
    3 &  0.856 $\pm$ 0.021 & 0.65 $\pm$ 0.06 & 11.1 & 0.854 $\pm$ 0.021 & 0.76 $\pm$ 0.07 & 5.6\\
    4 &  - & - & $> 12$ & 0.859 $\pm$ 0.020 & 0.73 $\pm$ 0.06 & 6.7\\
    5 &  - & - & $> 12$ & 0.866 $\pm$ 0.020 & 0.71 $\pm$ 0.06 & 8.1\\
    \hline \\
\end{tabular}
\caption{Average Dice scores, percentage of negative Jacobian's determinant, and memory usage for the original VoxelMorph architecture and the contracted architecture, for different number of cascades.}
\label{contracted}
\end{table*}

\subsubsection*{Reducing folding transformation}

Ideally, the transformation that aligns the moving image with the fixed image should be smooth and invertible. Non-invertible transformations often contain multiple regions of folding, which can be quantified by the percentage of negative determinant of the transformation's Jacobian. One way to reduce the amount of folding is to use diffeomorphic registration, although this method is generally associated with lower registration accuracy. A key factor in minimizing folding is the regularization parameter $\lambda$ in Equation \ref{loss}, which helps to ensure smooth deformation by minimizing the average local gradient. By carefully selecting the value of $\lambda$, it is possible to balance the need for smooth deformations with the accuracy of the registration. Table \ref{tab_jac} shows the percentage of negative Jacobian determinant of the transformation for different values of $\lambda$, as well as the average Dice score obtained. Based on these results, we adopted a regularization parameter of $\lambda=1$, which provides the best Dice score and significantly reduces the percentage of negative Jacobian values.

\begin{table}[!ht]
\centering
\begin{tabular}{p{1.5cm}>{\centering\arraybackslash}p{3cm}>{\centering\arraybackslash}p{3cm}}    
    \hline
    $\lambda$ & $\% |J_\varphi | < 0$ & Dice\\ 
    \hline
    $10^{-4}$ & 6.11 $\pm$ 0.07 & 0.849 $\pm$ 0.021\\
    $10^{-1}$ & 3.15 $\pm$ 0.06 & 0.859 $\pm$ 0.020\\
    1 & 0.71 $\pm$ 0.06 & 0.866 $\pm$ 0.020\\
    2 & 0.21 $\pm$ 0.03 & 0.852 $\pm$ 0.026\\
    \hline \\
\end{tabular}
\caption{Average Dice scores and percentage of negative Jacobian's determinant for different regularization parameters.}
\label{tab_jac}
\end{table}

\begin{figure}[!ht]
  \centering
  \includegraphics[width=.95\textwidth]{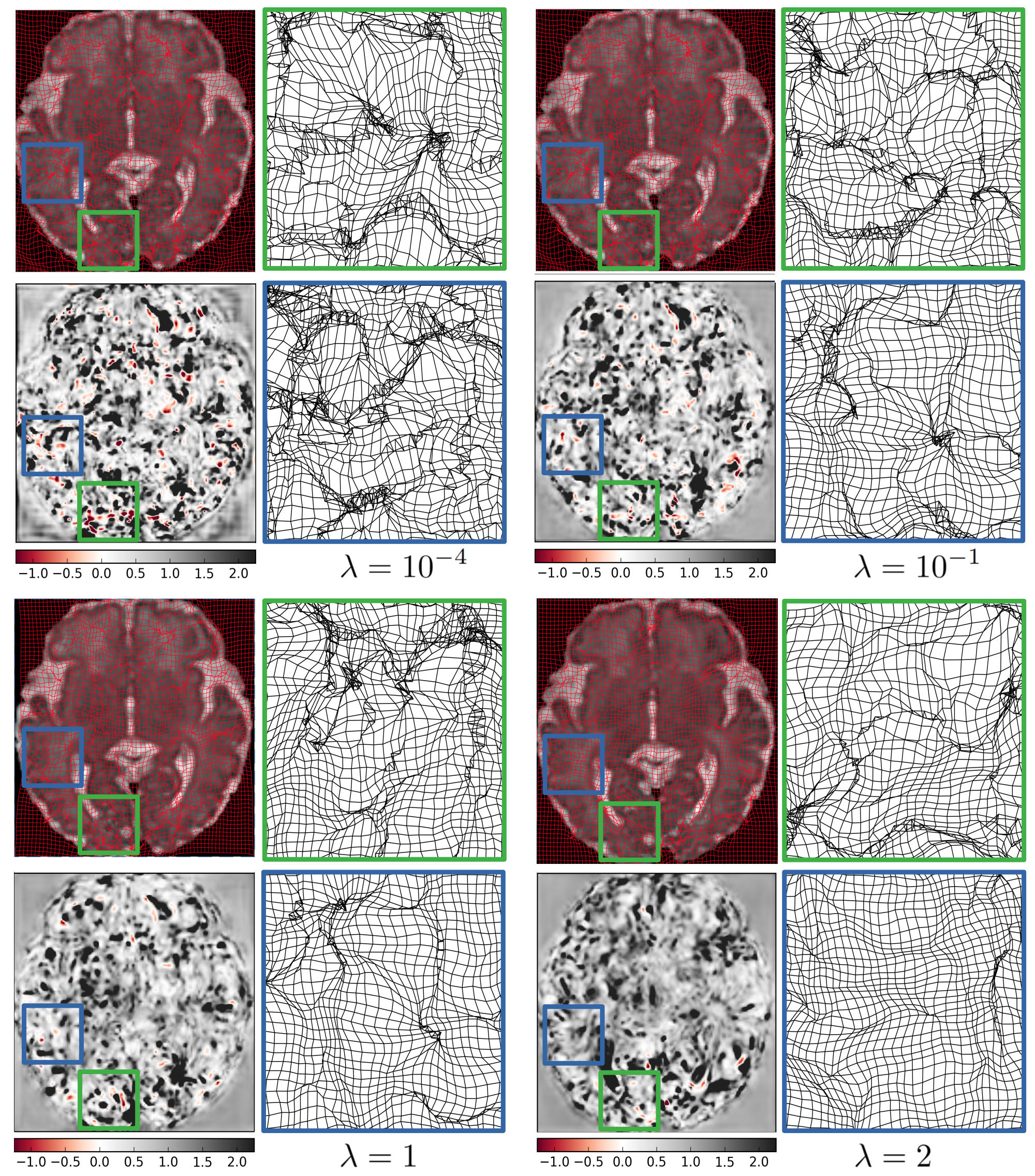}
  \caption{Example of a warped image with different regularization parameter $\lambda$, with the deformation grid superimposed, the corresponding Jacobian's determinant map and a zoom of the deformation grid on two regions.}
  \label{jdet_figure}
\end{figure}

Figure \ref{jdet_figure} illustrates the effect of the regularization parameter $\lambda$ on the smoothness of the deformation field. The figure includes a warped image with a grid overlay, a map of the determinant of the Jacobian, and a zoomed-in view of two folding regions for different values of the regularization parameter. It visually demonstrates that higher values of $\lambda$ lead to smoother deformation fields and fewer regions of voxel folding (i.e., regions with negative Jacobian's determinant).

\subsubsection*{Registration accuracy}

Following the cascaded registration method presented in Section \ref{casc_reg_sec}, we evaluated our registration model on two fetal brain MRI scans datasets. The quantitative evaluation is presented in Table \ref{table_reg}. It shows the average Dice score for all seven anatomical labels between the ground-truth labels of the fixed image and the propagated labels from the atlases. While the network was trained by maximizing the image similarity between the fixed image and the warped image with $\varphi$ using bilinear interpolation, the propagated labels are generated by warping the labels of the moving image with $\varphi$ using nearest neighbor interpolation. The results of our method are presented together with the results of VoxelMorph \cite{Voxelmorph} and TransMorph \cite{chen2022transmorph}. The Dice score after rigid registration is also given as reference. Two strategies of cascaded registration were evaluated: CasReg corresponds to the registration for which the moving image is successively warped by the cascaded networks without deformation field accumulation, while CasReg-acc is the strategy with deformation field accumulation, for which the accumulated deformation field is only applied to the moving image, as discussed in Section \ref{casc_reg_sec}. Although CasReg, with an average Dice of $0.837 \pm 0.021$, performs significantly better than VoxelMorph \cite{Voxelmorph} and TransMorph, with average Dice scores of $0.811 \pm 0.023$ and $0.807 \pm 0.23$, respectively, the accumulated strategy achieves the best results by a substantial margin, with a Dice score of $0.866 \pm 0.020$.

\begin{table*}[!ht]
\centering
\begin{tabular}{>{\centering\arraybackslash}p{2.3cm}>{\centering\arraybackslash}p{2.3cm}>{\centering\arraybackslash}p{2.3cm}>{\centering\arraybackslash}p{2.3cm}>{\centering\arraybackslash}p{2.3cm}>{\centering\arraybackslash}p{2.3cm}}
    \hline
    Label & Rigid & VoxelMorph & TransMorph &  CasReg & CasReg-acc \\
    \hline
    CSF &  0.516 $\pm$ 0.008 & 0.802 $\pm$ 0.007 & 0.796 $\pm$ 0.009 & 0.823 $\pm$ 0.007 & \bf{0.855 $\pm$ 0.008}\\
    Grey Matter & 0.501 $\pm$ 0.003 & 0.724 $\pm$ 0.009 & 0.725 $\pm$ 0.011 & 0.754 $\pm$ 0.009 & \bf{0.781 $\pm$ 0.006}\\
    White Matter & 0.663 $\pm$ 0.003 & 0.782 $\pm$ 0.005 & 0.780 $\pm$ 0.007 & 0.821 $\pm$ 0.006 & \bf{0.850 $\pm$ 0.004}\\
    ventricles & 0.495 $\pm$ 0.011 & 0.761 $\pm$ 0.017 & 0.751 $\pm$ 0.016 & 0.787 $\pm$ 0.013 & \bf{0.812 $\pm$ 0.012}\\
    Cerebellum & 0.784 $\pm$ 0.010 & 0.881 $\pm$ 0.009 & 0.873 $\pm$ 0.014 & 0.899 $\pm$ 0.007 & \bf{0.932 $\pm$ 0.005}\\
    Thalamus & 0.812 $\pm$ 0.004 & 0.871 $\pm$ 0.008 & 0.871 $\pm$ 0.007 & 0.884 $\pm$ 0.007 & \bf{0.915 $\pm$ 0.004}\\
    Brain Stem & 0.767 $\pm$ 0.005 & 0.856 $\pm$ 0.011 & 0.853 $\pm$ 0.012 & 0.889 $\pm$ 0.009 & \bf{0.919 $\pm$ 0.004}\\
    \hline
    Average & 0.648 $\pm$ 0.054 & 0.811 $\pm$ 0.023 & 0.807 $\pm$ 0.023 & 0.837 $\pm$ 0.021 & \bf{0.866 $\pm$ 0.020}\\ 
    \hline
\end{tabular}
\caption{Average Dice scores obtained using ANTs rigid registration \cite{avants2011reproducible}, VoxelMorph \cite{Voxelmorph}, TransMorph\cite{chen2022transmorph} and the proposed method, with and without deformation field accumulation.}
\label{table_reg}
\end{table*}

\begin{figure*}[!ht]
  \centering
  \includegraphics[width=.77\textwidth]{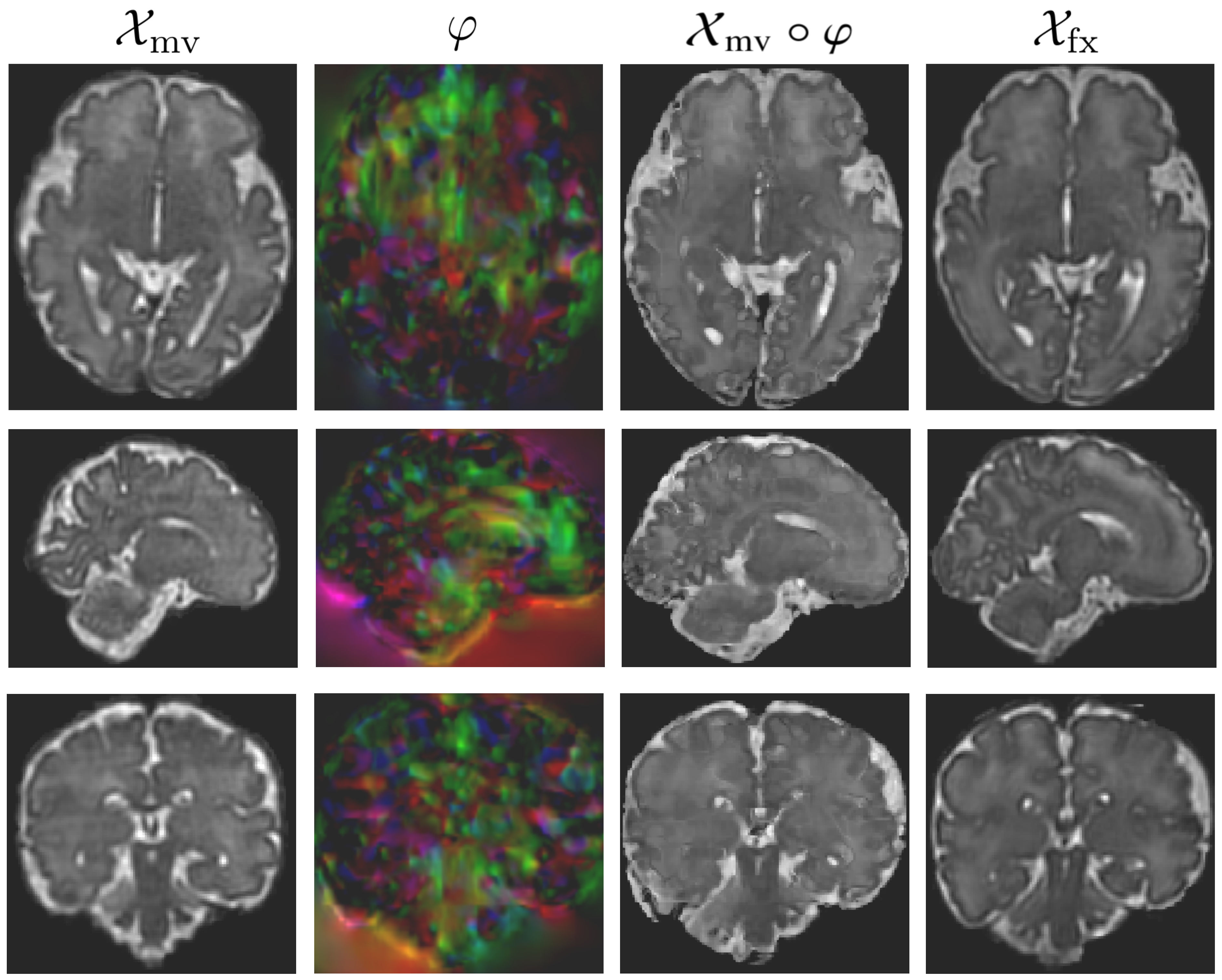}\hfill
  \includegraphics[width=.77\textwidth]{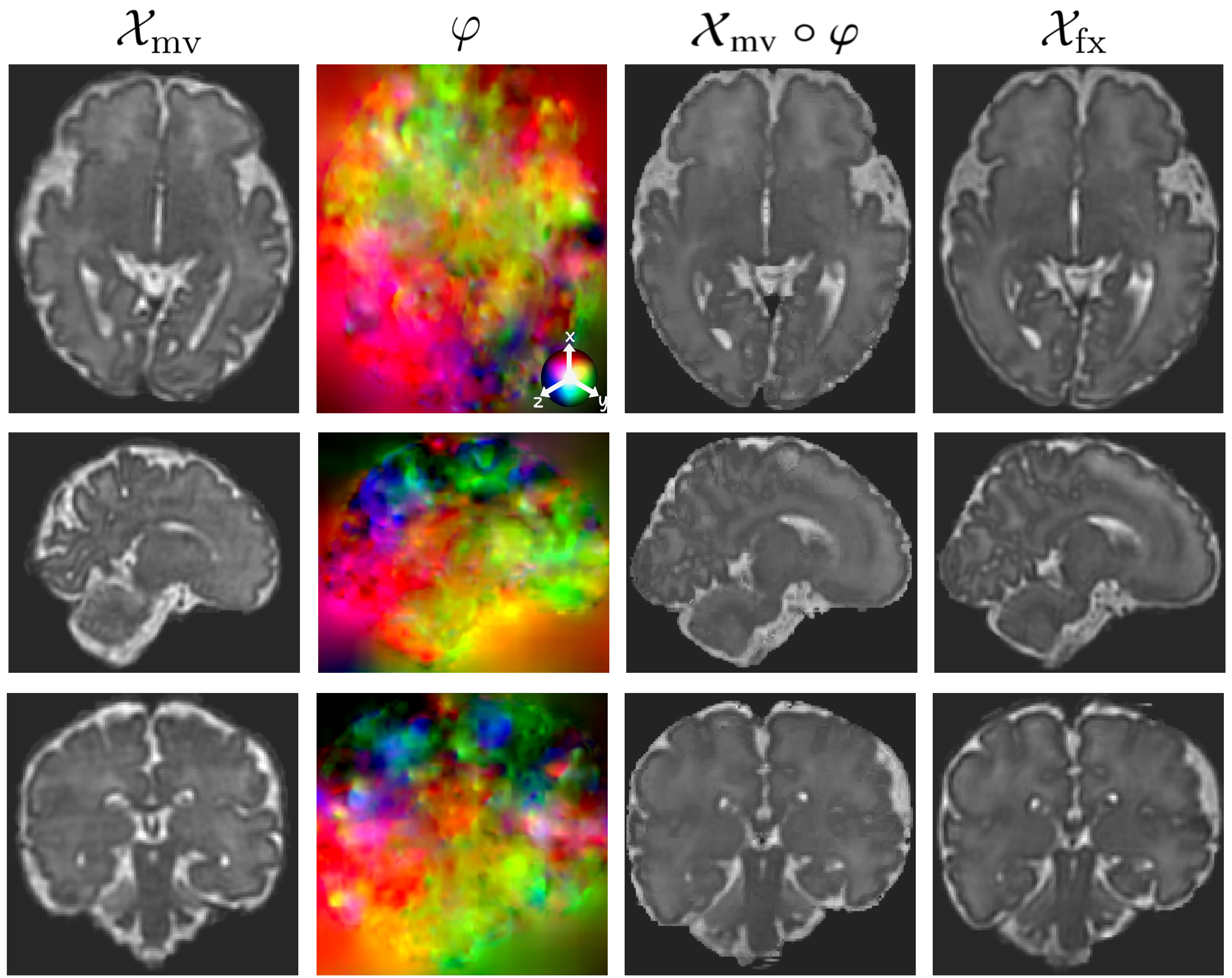}
  \caption{Comparison of the qualitative results of VoxelMorph (top) and our proposed method (bottom). The axial, sagittal and coronal views of the moving, warped and fixed images are shown, as well as the corresponding deformation field.}
  \label{fig_flows}
\end{figure*}

A comparison of the qualitative results of our method with VoxelMorph \cite{Voxelmorph} is shown in Figure \ref{fig_flows}. It clearly establishes that the cascaded registration produces better and more realistic results. The appearance of the deformation field is also substantially different for both methods. This is due to the fact that the cascaded approach computes the deformation fields at different scales, rather than a single, global deformation field. Hence, the final deformation field is able to capture more accurately the local variations in the shape and structure of the fixed image.

\subsection{Multi-atlas segmentation}

\subsubsection*{Atlas selection}\label{res_atlas_selection}

As mentioned in Section \ref{atlas_selection}, the selection of atlases for MAS is an important factor that can affect the accuracy of the segmentation. By carefully choosing atlases that are similar to the target image, it is possible to improve the accuracy of the segmentation by eliminating atlases that are not relevant. Traditionally, atlas selection has been performed after registration, which can be computationally intensive and time-consuming. However, the use of DL methods has greatly reduced the computation time required for registration, allowing for the selection of atlases based on similarity metrics to be performed efficiently. With our approach, a pairwise registration is completed in approximately 0.2 seconds, which allows us to register a large number of atlases with the target image and select the most suitable ones without increase exceedingly the overall computation time.

\begin{figure*}[!ht]
\centering
\includegraphics[width=.3\textwidth]{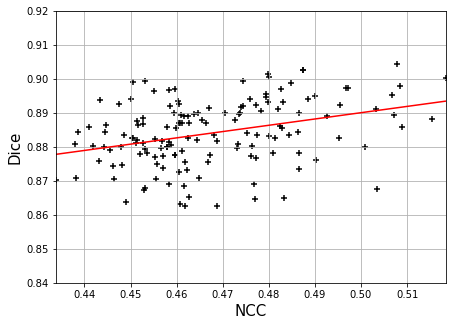}\hfill
\includegraphics[width=.3\textwidth]{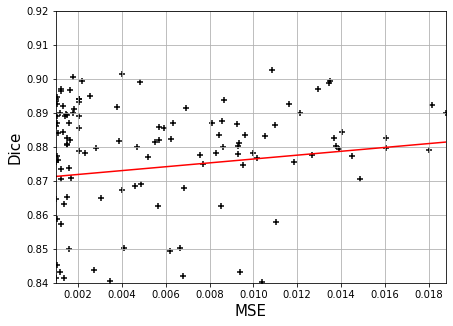}\hfill
\includegraphics[width=.3\textwidth]{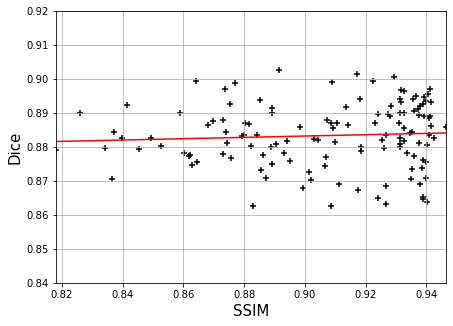}
\caption{Dice score as a function of the similarity metrics for Normalized Cross Correlation, Mean Square Error, and Structural Similarity Index Measure, for the IMPACT dataset.}
\label{fig:atlas_sel}
\end{figure*}

We examined the effect of atlas selection using three different similarity metrics: NCC, MSE and SSIM. To evaluate the performance of each metric, we calculated the Dice score of the propagated labels and plotted the results in Figure \ref{fig:atlas_sel}. The figure shows that there is a stronger correlation between higher values of NCC and higher Dice scores. Therefore, we adopted NCC as the similarity metric for atlas selection.

\begin{figure}[!t]
  \centering
  \includegraphics[width=.45\textwidth]{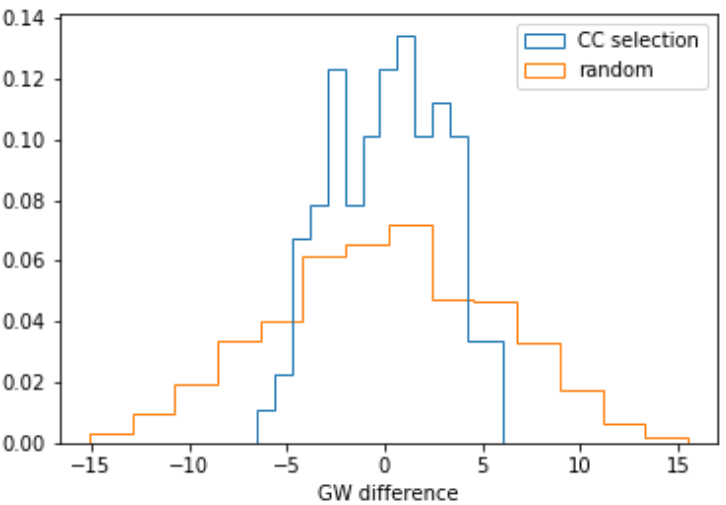}
  \caption{Histogram of the gestational age difference between the target image and the selected atlases (orange), compared to a random selection (blue).}
  \label{gw_diff_cc_sel}
\end{figure}

In addition, we conducted an experiment on the FeTA dataset to further demonstrate the effectiveness of NCC-based atlas selection. The FeTA dataset includes fetuses at various gestational ages, which provides a good opportunity to test the ability of our method to select appropriate atlases, based on the gestational age deviation with the target image. Figure \ref{gw_diff_cc_sel} compares the distribution of gestational age difference between the NNC-selected atlases and the target image to the distribution obtained without selection. Each target image was registered to the other images of the test set, and ten atlases were selected based on NCC in the first case, whereas ten atlases were randomly chosen in the second case. The figure shows that the NCC-selection significantly reduces the age difference compared to the random selection. The selected atlases stand within a range of $\pm 5$ gestational week difference from the target image. This result indicates that our atlas selection method not only leads to better alignment of the propagated labels, but also favors atlases that are relatively close in terms of gestational age.

\subsubsection*{Label fusion}

\begin{figure*}[!ht]
  \centering
  \includegraphics[width=\textwidth]{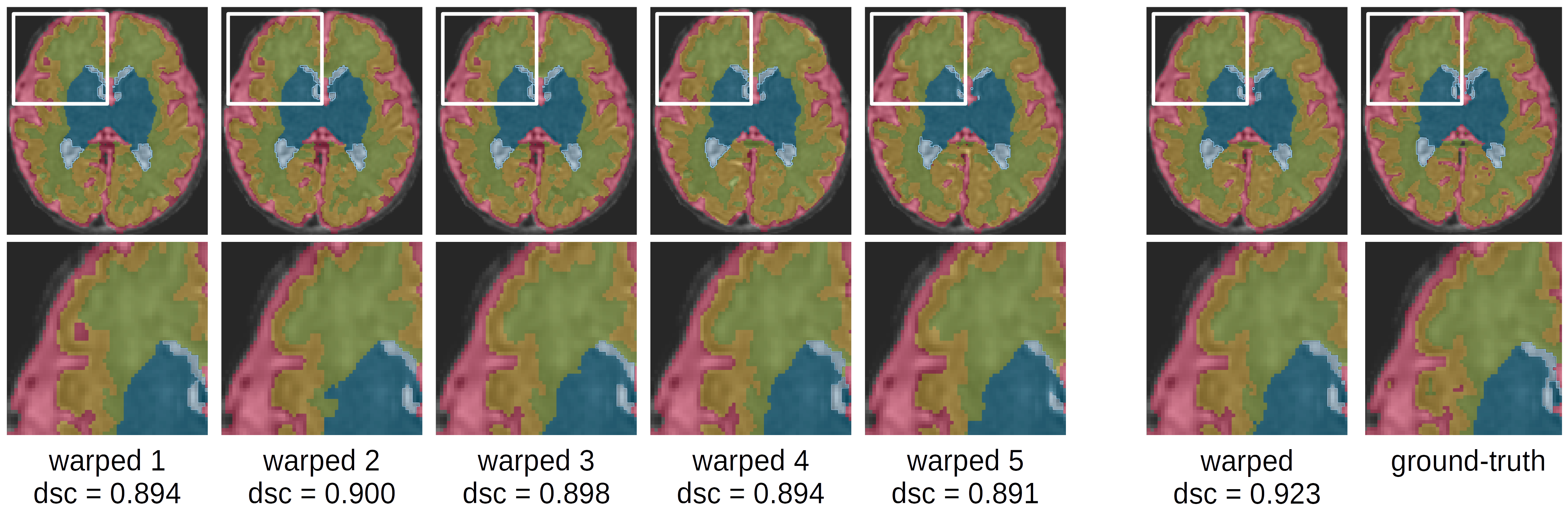}
  \caption{Example of the label selection by local weighted voting. The average Dice scores show the improvement between the individual warped labels and the refined one.}
  \label{labelsel}
\end{figure*}

The label fusion strategy we adopted is based on LWV, as discussed in Section \ref{label_fusion}. LWV, compared to MV, represents a marginal improvement, with an average Dice score of $0.926 \pm 0.012$ versus $0.915 \pm 0.012$, respectively. Figure \ref{labelsel} shows an example of propagated labels and the refined segmentation using LWV. In this example, the average Dice score over all seven anatomical labels ranges from $0.891$ to $0.900$ for the propagated labels, while the refined labels generated with LWV obtains a Dice score of $0.923$.

\subsubsection*{Segmentation accuracy} 

In this section, we present the segmentation results of our MAS method using cascaded registration. We provide a comparison with two classical MAS approaches \cite{ballester2022automatic, payette2021automatic} for the IMPACT dataset. Additionally, we compare our segmentation results with nnU-Net \cite{isensee2021nnu} for both IMPACT and FeTa datases. The procedure for producing the refined segmentation involves first to register each of the images of the test set to the target image using the cascaded registration approach with the deformation field's accumulation strategy (Section \ref{casc_reg_sec}), then to select the best aligned images based on the NCC of the warped images with the target image and propagate the corresponding labels (Section \ref{atlas_selection}), and finally to combine the propagated labels using a local voting strategy based on the local NCC (Section \ref{label_fusion}).

Table \ref{tab_seg} presents the segmentation results on the 20 fetal brain MRI scans of the IMPACT test set, nnU-Net \cite{isensee2021nnu} when trained and tested on the same sets of images, and two classical MAS methods that are the dHCP pipeline for neonatal brain segmentation \cite{makropoulos2018developing} and the perinatal brain segmentation pipeline \cite{ballester2022automatic}. It shows that our method obtains similar results than nnU-Net for all anatomical labels, achieving an average Dice score of $0.926 \pm 0.012$ versus $0.920 \pm 0.014$ for nnU-Net. The tables also demonstrate the efficiency of our method compared to classical MAS pipelines for fetal and neonatal brain MRI scans, which obtain an average Dice score significantly lower with $0.87 \pm 0.02$ for the perinatal brain segmentation pipeline and $0.81 \pm 0.03$ for the dHCP pipeline for neonatal segmentation. The other advantage of our MAS approach when compared to the classical methods is the drastically reduced computational time. For the same single segmentation task, classical approaches require from 30 minutes up to one hour, whereas our method segments the image in approximately 60 seconds, in the same range as nnU-Net.

\begin{table*}[!ht]
\centering
\begin{tabular}{>{\centering\arraybackslash}p{2.5cm}>{\centering\arraybackslash}p{2.5cm}>{\centering\arraybackslash}p{2.5cm}>{\centering\arraybackslash}p{2.5cm}>{\centering\arraybackslash}p{2.5cm}}
    \hline
    Label & Ours & nnU-Net & PP & dHCP \\
    \hline
    CSF & \bf{0.923 $\pm$ 0.006} & 0.897 $\pm$ 0.011 & 0.83 $\pm$ 0.04 & 0.79 $\pm$ 0.02\\
    Grey Matter & \bf{0.877 $\pm$ 0.006} & 0.871 $\pm$ 0.012 & 0.85 $\pm$ 0.03 & 0.76 $\pm$ 0.05\\
    White Matter & 0.915 $\pm$ 0.006 & \bf{0.919 $\pm$ 0.006} & 0.90 $\pm$ 0.02 & 0.85 $\pm$ 0.02\\
    Ventricles & \bf{0.902 $\pm$ 0.005} & 0.879 $\pm$ 0.011 & 0.73 $\pm$ 0.05 & 0.66 $\pm$ 0.07\\
    Cerebellum & 0.964 $\pm$ 0.004 & \bf{0.965 $\pm$ 0.004} & 0.93 $\pm$ 0.02 & 0.89 $\pm$ 0.03\\
    Thalamus & 0.950 $\pm$ 0.002 &\bf{0.957 $\pm$ 0.004} & 0.92 $\pm$ 0.02 & 0.88 $\pm$ 0.03\\
    Brain Stem & \bf{0.955 $\pm$ 0.003} & 0.953 $\pm$ 0.006 & 0.91 $\pm$ 0.01 & 0.90 $\pm$ 0.02\\
    \hline
    Average & \bf{0.926 $\pm$ 0.012} & 0.920 $\pm$ 0.014 & 0.87 $\pm$ 0.02 & 0.81 $\pm$ 0.03\\
    \hline
    Time & ~60s & ~80s & $>30$min & $> 1$h\\
    \hline \\
\end{tabular}
\caption{Average Dice scores obtained using MAS, nnU-Net \cite{isensee2021nnu}, the perinatal brain segmentation pipeline PP \cite{ballester2022automatic}, and the dHCP pipeline for neonatal segmentation \cite{makropoulos2018developing}, on the IMPACT dataset \cite{ballester2022automatic}.}
\label{tab_seg}
\end{table*}

Table \ref{tab_seg_feta} showcases the segmentation outcomes on 20 fetal brain MRI scans from the FeTa test set \cite{payette2021automatic}, obtained using our MAS method and nnU-Net \cite{isensee2021nnu}. In this case, nnU-Net outperforms our approach for 5 out of the 7 anatomical labels, achieving a higher average Dice score of $0.866 \pm 0.013$ compared to $0.847 \pm 0.008$ for our method. The comparatively lower performance on the FeTa dataset can be attributed to several factors. Firstly, the FeTa dataset exhibits a greater morphological variability due to the inclusion of fetuses with pathologies that impact the shapes of certain brain regions, such as ventricles in the case of VM. Moreover, the dataset encompasses a broader range of gestational ages, from 20 to 35 weeks, as opposed to the 32 to 37-week range in the IMPACT dataset. Secondly, the FeTa dataset is characterized by a higher degree of heterogeneity as the fetal brain MRI scans were acquired from multiple centers using different scanners and 3D SR techniques, resulting in diverse image appearances within the dataset. This increased variability in morphology and appearance complicates the task of establishing correspondences between image pairs, thereby posing challenges for registration-based segmentation.

The segmentation results of our proposed method on the two fetal brain MRI datasets were generally satisfactory. Although nnU-Net outperformed our method on the FeTa dataset, a notable advantage of our approach is its ability to effectively utilize a limited number of annotated images for MAS, without the need for labeled data during network training. This aspect can lead to considerable savings in terms of time and resources, particularly in scenarios where annotated data is scarce. Moreover, our method offers the benefit of enforcing spatial and topological consistency in the segmentation results, as opposed to deep learning-based segmentation methods. This could be an important advantage when dealing with medical images, where preserving accurate anatomical relationships and structure is crucial for reliable interpretation and analysis.

\begin{table}[!ht]
\centering
\begin{tabular}{>{\centering\arraybackslash}p{2.5cm}>{\centering\arraybackslash}p{2.5cm}>{\centering\arraybackslash}p{2.5cm}}
    \hline
    Label & Ours & nnU-Net\\
    \hline
    CSF & \bf{0.903 $\pm$ 0.006} & 0.878 $\pm$ 0.012\\
    Grey Matter & 0.742 $\pm$ 0.010 & \bf{0.762 $\pm$ 0.021}\\
    White Matter & 0.913 $\pm$ 0.003 & \bf{0.914 $\pm$ 0.006}\\
    Ventricles & 0.816 $\pm$ 0.023 & \bf{0.902 $\pm$ 0.014}\\
    Cerebellum & \bf{0.902 $\pm$ 0.005} & 0.897 $\pm$ 0.008\\
    Thalamus & 0.837 $\pm$ 0.016 & \bf{0.866 $\pm$ 0.012}\\
    Brain Stem & 0.814 $\pm$ 0.017 & \bf{0.841 $\pm$ 0.013}\\
    \hline
    Average & 0.847 $\pm$ 0.008 & \bf{0.866 $\pm$ 0.013}\\
    \hline
\end{tabular}
\caption{Average Dice scores obtained using MAS, nnU-Net \cite{isensee2021nnu}, on the FeTa dataset \cite{payette2021automatic}.}
\label{tab_seg_feta}
\end{table}

\section{Discussion and conclusions}\label{conclusion}

In this work, we introduce a registration model based on cascaded networks for MAS of fetal brain MRI scans. Our model decomposes the registration field into a series of simpler transformations that operate at different spatial scales, allowing for more efficient and accurate alignment of the atlases with the target image. One key aspect of our approach is the accumulation of the deformation fields produced by the cascaded network, which avoids the loss of information that can occur with successive warping and interpolation.

An interesting property of our cascaded networks is their ability to capture deformations at different spatial resolutions. Indeed, each network focuses on a different aspect of the registration, from capturing global deformations in the earlier stages to refining local deformations and capturing high-frequency details in the later stages. One reason cascaded networks can produce deformation fields at different spatial resolutions is the hierarchical nature of their processing. In each stage of the cascade, the network receives input from the previous stage, allowing it to build upon and refine the transformation. This hierarchical processing helps the networks to capture more complex and non-linear relationships between the images, which are often present in medical image registration problems. Furthermore, the cascaded approach enables the networks to learn different levels of image abstraction. The first network in the cascade focuses on the coarsest representation, which corresponds to low spatial resolution. As we move through the cascade, each network refines the registration, moving towards higher spatial resolutions. This gradual increase in spatial resolution allows the networks to capture both global and local deformations more effectively.

To evaluate the performance of our method, we conducted experiments on two fetal brain MRI datasets and compared the results to state-of-the-art methods \cite{Voxelmorph,chen2022transmorph}. Our results show that our approach outperforms these methods in both quantitative and qualitative measures. In addition, the derived MAS method achieves performance similar to one of the most robust state-of-the-art segmentation methods \cite{isensee2021nnu}, despite not requiring any labelled training data. This novel MAS methodology, utilizing a selected subset of annotated data in lieu of traditional atlases, offers a viable alternative for contexts where the amount of annotated data is limited.

One of the significant advantages of the MAS approach over DL-based segmentation methods, is its inherent ability to enforce spatial and topological consistency. This is particularly important in medical image segmentation, where maintaining the anatomical structures and their relationships is crucial for accurate diagnosis and analysis. Furthermore, our MAS method can adapt to different annotation standards and requirements and can be used in multi-center studies, where different centers may have their own specific annotation protocols. It would be therefore possible to incorporate data from multiple centers without the need to standardize the annotation process, potentially saving time and resources.

We look forward to exploring further improvements, such as the local regularization of the deformation field, through the use of locally weighted constraints based on biomechanical models, for instance. Further investigations will be also directed towards the analysis of longitudinal data from fetal and neonatal brains. In this context, precise registration can be useful to improve biomechanical models of perinatal brain growth. Additionally, we aim to investigate the effectiveness of our approach in other medical imaging applications where limited annotated training data is a common challenge.

\section{Acknowledgements}

This publication is part of the project PCI2021-122044-2A, funded by the project ERA-NET NEURON Cofund2, by MCIN/AEI/10.13039/501100011033/, FIS\_AC21-AC21\_2/00016-PI047479 and by the European Union “NextGenerationEU”/PRTR.
G. Piella is supported by ICREA under the ICREA Academia programme. 

We also thank the study participants for their personal time and commitment to the IMPACT BCN Trial, and all the medical staff, residents, midwives, nurses, MR platform, and researchers of BCNatal especially Annachiara Basso, MD and Kilian Vellvé, MD for their support in the MR data collection. IMPACT BCN Trial was partially funded by a grant from “la Caixa” Foundation (LCF/PR/ GN18/10310003); Cerebra Foundation for the Brain Injured Child (Carmarthen, Wales, UK); ASISA Foundation; AGAUR under grant 2017 SGR No. 1531 and Instituto de Salud Carlos III (ISCIII), PI18/00073, co-funded by the European Union. A. Nakaki has received the support of a fellowship from la Caixa Foundation under grant number LCF/BQ/DR19/11740018. F.Crovetto reports a personal fee from Centro de Investigaciones Biomédicas en Red sobre Enfermedades Raras (CIBERER).


\bibliographystyle{plain}
\bibliography{main}

\end{document}